\documentclass[conference]{IEEEtran}
\usepackage[utf8]{inputenc}
\usepackage{graphicx}
\usepackage{hyperref}

\usepackage{algorithm}
\usepackage{algorithmic}

\usepackage{amsmath}
\usepackage{amsfonts}
\usepackage{amssymb}

\usepackage{graphicx}
\usepackage{wrapfig}
\usepackage{hyperref}
\usepackage{xspace}
\usepackage{color}
\usepackage{parskip}
\usepackage{multicol}
\usepackage{mathptmx}
\usepackage{fancyhdr}
\usepackage{eurosym}
\usepackage{wasysym}
\usepackage{bbm}

\def\mR{{\mathbb{R}}}

\def\mRplus{\mR_{+}}
\def\mRplus0{\mR_{+,0}}

\def\SO#1{\mbox{SO}( #1 )}
\def\axis{q}
\def\rotx#1{\hbox{R}_x({#1})}
\def\roty#1{\hbox{R}_y({#1})}
\def\rotz#1{\hbox{R}_z({#1})}
\def\transx#1{\hbox{T}_x({#1})}
\def\transy#1{\hbox{T}_y({#1})}
\def\transz#1{\hbox{T}_z({#1})}
\def\trans#1#2#3{\hbox{Trans}({#1}, #2, #3)}
\def\tcp{\ifmmode \mbox{\sf TCP} \else \mbox{\sf TCP}\xspace\fi}
\def\TCP{\tcp}
\def\tool{\mbox{\sf TOOL}}
\def\atan2#1#2{\mbox{atan2}({#1}, {#2})}
\def\world{{\sf W}}
\def\workspace{{\cal W}}
\def\wcp{\ifmmode \mbox{\sf WCP} \else \mbox{\sf WCP}\xspace\fi}
\def\WCP{\wcp}
\def\equote#1{``#1''}

\def\rotx#1{\hbox{R}_x\left({#1}\right)}
\def\roty#1{\hbox{R}_y\left({#1}\right)}
\def\rotz#1{\hbox{R}_z\left({#1}\right)}

\begin{document}

\title{Optimization of Robot Tasks with Cartesian Degrees of Freedom using Virtual Joints}

\author{
\IEEEauthorblockN{Martin Wei\ss}
\IEEEauthorblockN{Martin.Weiss@oth-regensburg.de}
\IEEEauthorblockA{Fakult\"at für Informatik und Mathematik\\
Ostbayerische Technische Hochschule Regensburg\\
93049 Regensburg}
}

\def\frames{{\cal {F}}}
\def\SO#1{\mbox{SO}( #1 )}
\def\axis{q}
\def\rotx#1{\hbox{R}_x({#1})}
\def\roty#1{\hbox{R}_y({#1})}
\def\rotz#1{\hbox{R}_z({#1})}
\def\transx#1{\hbox{T}_x({#1})}
\def\transy#1{\hbox{T}_y({#1})}
\def\transz#1{\hbox{T}_z({#1})}
\def\trans#1#2#3{\hbox{Trans}({#1}, #2, #3)}
\def\tcp{\ifmmode \mbox{\sf TCP} \else \mbox{\sf TCP}\xspace\fi}
\def\TCP{\tcp}
\def\tool{\mbox{\sf TOOL}}
\def\atan2#1#2{\mbox{atan2}({#1}, {#2})}
\def\world{{\sf W}}
\def\workspace{{\cal W}}
\def\wcp{\ifmmode \mbox{\sf WCP} \else \mbox{\sf WCP}\xspace\fi}
\def\WCP{\wcp}
\def\equote#1{``#1''}

\renewcommand{\algorithmicrequire}{\textbf{Input:}}
\renewcommand{\algorithmicensure}{\textbf{Output:}}

\maketitle              %

\begin{abstract}
A common task in robotics is unloading identical goods from a tray with rectangular
grid structure. This naturally leads to the idea of programming the process at one grid
position only and translating the motion to the other grid points, saving teaching time.
However this approach usually fails because of joint limits or singularities of the robot. 
If the task description has some redundancies, e.g. the objects are cylinders where 
one orientation angle is free for the gripping process, the motion may be modified
to avoid workspace problems.
We present a mathematical algorithm that allows the automatic generation of robot
programs for pick-and-place applications with structured positions when the
workpieces have some symmetry, resulting in a Cartesian degree of freedom for 
the process. The optimization uses the idea of a virtual joint which measures the 
distance of the desired \TCP to the workspace such that the nonlinear optimization
method is not bothered with unreachable positions. Combined with smoothed versions 
of the functions in the nonlinear program higher order algorithms
can be used, with theoretical justification superior to many ad-hoc approaches used
so far.

\end{abstract}
\section{Problem Statement}

\begin{figure}
\centerline{\includegraphics[height=35mm]{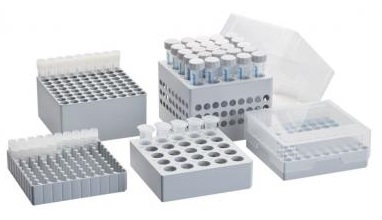}}
\caption{Storage boxes} 
\label{fig:StorageBoxes}
\end{figure}

We consider the following task: A robot should unload a storage box with a chess-board
like structure containing $B_x\times B_y$ identical workpieces at positions $P_{kl}$, 
$k=1, \ldots, B_x$, $l = 1, \ldots, B_y$, counted in the 
coordinate directions of the frame $C\in\frames$ 
(where $\frames$ denotes the set of all frames $\mR^3\times\SO 3$) associated with the 
box at distances $D_x$ and $D_y$. Think of test-tubes in medicine or small parts 
in general production, as in Figure \ref{fig:StorageBoxes}. 
The cell setup is considered fixed, also the placement of the box 
in the cell cannot be chosen.

Usually a pick-and-place operation is programmed at one corner only, the other
position commands are computed from this corner position and the indices and distances. 
In this paper we consider the simplest case of a linear motion from the workpiece positions
$P_{kl}$ to a position $P_{kl}+ {[0, 0, \Delta z]} t$ some safe distance $\Delta z$ 
above the grid from where the object can be moved with PTP motions which are considered
\equote{simple} in this paper so there is no need for optimization.

We consider a standard 6-axis kinematics with central wrist where up to 8 discrete 
solutions of the backward transform can be calculated analytically in non-singular 
configurations, one of which is selected by some configuration bits in the application
program. We identify these 8 configurations with an integer 
$s\in\{0, \ldots, 7\}$.
However it is difficult for the user to assess whether all positions are reachable because 
of nonlinearity, singularities, axis limits, cabling restricting the axes, and so on. 
Testing the corners is a heuristic that works in many cases but there is
no guarantee, so one has to run time-consuming tests. When the process needs to work
on an object from different sides or with different orientations the situation is even
more complicated. So the user would like to have an algorithm that determines a feasible
object frame $C$ near some initial guess $C_0$, maybe additionally optimizing one of the
many known manipulability measures, see \cite{Merlet}, \cite{Yoshikawa}.

\begin{figure}
\centerline{\includegraphics[height=50mm]{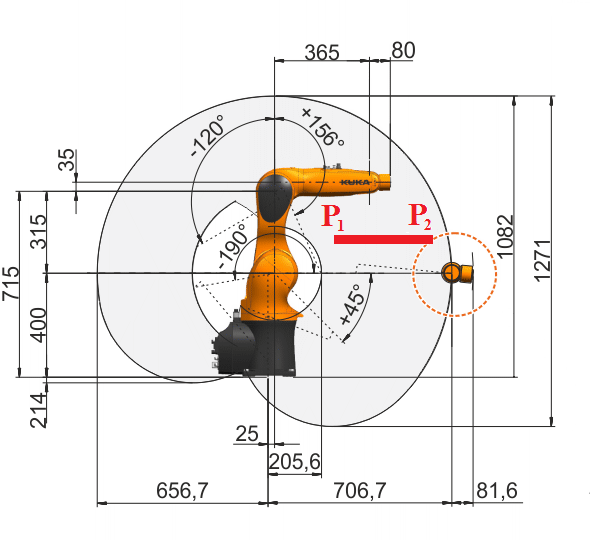}}
\caption{Cross section of workspace with storage box} 
\label{fig:AgilusWorkspace}
\end{figure}

Figure \ref{fig:AgilusWorkspace} shows a workspace cross section of 
a {KUKA Agilus KR 6R700 sixx} industrial robot as in the manufacturer's documentation
\cite{KUKASpezAgilus} with a storage box with 2 objects only drawn as a red bar, 
and $P_{1}\equiv P_{1,1}\in\mR^3$ and $P_2\equiv P_{1,2}\in\mR^3$ as 
corners\footnote{Source for storage box picture: \url{https://www.eppendorf.com}}. 
Both points are inside the Cartesian workspace of the robot but $P_2$ is so close 
to the workspace boundary such that the flange cannot be oriented parallel to the box
to lift the object along the $z$ direction. For a complete specification of the robot
pose also the orientation has to be specified (plus some configuration bits). In our 
application the user wants the tool direction to be perpendicular to the grid box, so 
one degree of freedom - the orientation around the tool direction - remains for 
optimization. None of the state-of-the art robot programming languages offer constructs
that leave one degree of freedom for optimization: The expert mode of 
KUKA's KRL for example allows to leave out some components of a Cartesian point (also 
of axis motions), like \verb+PTP {X 450, Y 0, Z 300, A 90, B 0}+ where the \verb+C+
 component
of the orientation is missing, but this is only a short hand writing for 
\equote{keep the current {\tt C} component}. So essentially all degrees of freedom are
specified. So the user has to specify a degree of freedom in a - usually - suboptimal 
way although it has no meaning for the process, and may even lead to unreachable 
poses.

Our goal is to find admissible motions for all grid points given one sample motion,
where the process degree of freedom is determined by an optimizer to give admissible
motions at all grid points. In order to use the standard programming languages these
motions are specified by all degrees of freedom: 5 given by the grid point and the 
sample motion, one determined by the optimizer.

One main difficultiy arises: 
It is easy to check in a program whether a given frame $C$ leads to
reachable positions or not like $P_{1}$ in the figure, 
or unreachable positions like $P_{2}$, 
but it is difficult for a nonlinear optimizer to determine a
direction that leads to a ''{more feasible}'' situation, starting from
an infeasible one: Feasibility is a binary
decision; the backward transformation 
will usually issue an error only, and abort.

Our idea is to introduce a virtual joint as a {\em slack variable} in 
terms of nonlinear programming (see \cite{Nocedal}) into the optimization
problem that measures the distance of a position from feasibility; . This variable 
therefore has an intuitive geometric interpretation. 
This approach has already been applied in \cite{WeissARK} to the optimal placement
of an object in the robot workspace when there are no redundant degrees of 
freedom in the process.

Our approach has some similarity to the introduction of virtual axes 
for singularity avoidance in 
\cite{Reiter} or \cite{Leontjevs}. However we do not introduce a rotational 
joint to reduce velocities near singularities but rather use a prismatic joint to
enlargen the mathematical workspace in the optimization process. In combination 
with a smooting operation we can use standard optimization algorithms which require
differentiability of order 1 like all algorithms based on gradient descent, 
or order 2 like Sequential Quadratic Programming (SQP), cf.~\cite{Nocedal}.

The paper is organzied as follows: In Section II we describe the idea of a virtual
axis in the kinematics. In Section III we state the optimization problem. 
Section IV shows numerical results, leading to the conclusion with directions
for further research in Section V.

\section{Virtual Axis Approach}

For ease of exposition we choose a 6R robot resembling the well known Puma 560 
or the KUKA Agilus but with more zeros in the parameters. We could extend all formulae 
to similar 6R real industrial robots. We use the DH convention 
$$
	\rotz {\axis_i} \cdot \transz {d_i} \cdot \transx {a_i} \cdot \rotx{\alpha_i} =: 
	\rotz {\axis_i} \cdot B_i =: 
	A_i(\axis_i) 
$$
to get wrist centre point \wcp and tool centre point \tcp
\begin{eqnarray*}
   \wcp(\axis) &=& A_1(\axis_1)\cdot  A_2(\axis_2)\cdot  A_3(\axis_3) \cdot 
		A_4(\axis_4) \cdot A_5(\axis_5) \cdot A_6(\axis_6) 
\\
	\tcp(\axis) &=& \wcp(\axis) \cdot \tool
\end{eqnarray*}
expressed relative to the world coordinate system chosen as the axis 1 coordinate 
system. 
\begin{table}
  \caption{DH parameters for original robot}
  \label{fig:DHparameters}
\centering
  \begin{tabular}{c|c|c|c|c||c}
        $i$ \   & \ $\theta_i$ \ & \ $d_i$ \   &  \  $a_i$ \ & \ $\alpha_i$ \ & \ type\\ 
	\hline
         1   & $q_1$ & 0   & 0  &  $\frac \pi 2$ & R\\
         2   & $q_2$ &  0 &   $l_{23}$ & 0 & R\\
         3   & $q_3$ &  0 & 0 & $-\frac \pi 2$ & R\\
         4   & $q_4$ &  $l_{35}$ & 0 & $\frac \pi 2$ & R\\
         5   & $q_5$ &  0 & 0  & $-\frac \pi 2$ & R\\
         6   & $q_6$ &  0   & 0 & 0 & R
  \end{tabular}
\end{table}
\begin{figure}
\centerline{
\includegraphics[width=60mm]{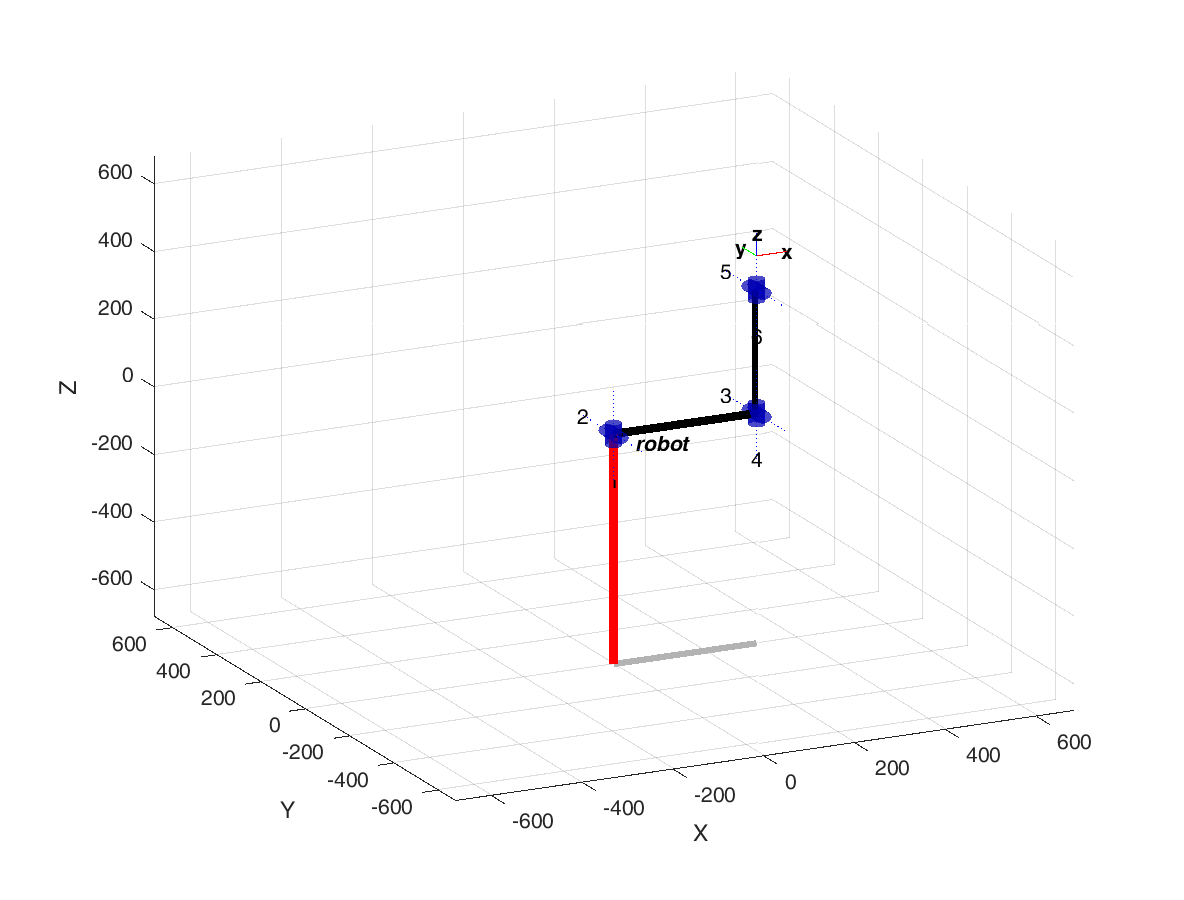}
}
  \caption{Reference position  $\axis=0$ for original robot}
  \label{fig:ZeroPositionparameters}
\end{figure}
Table \ref{fig:DHparameters} and Figure \ref{fig:ZeroPositionparameters} 
 show the robot data and the reference position. 
We assume joint limits
$-\pi \leq \axis_{\min,i} \leq \axis_i \leq \axis_{\max,i} \leq \pi$, $i=1, \ldots 6$. 

Infeasibility of the backward transform for a given frame $F$ and configuration
$s$ may arise from two reasons with different severity: 
First, the \wcp may be to far from the robot such that
the triangle construction for $\axis_3$ fails. There is no remedy in this case. 
Second, even if axis values $\axis$ exist such that $\tcp(\axis) = F$, these might 
violate the joint limits: $\axis_i\not\in [\axis_{\min,i}, \axis_{\max,i}]$ for some $i$. 
This is no obstacle during the optimization process, only for a solution. 
So the second problem can be fixed by dropping the joint limits and allowing
$\axis_i \in (-\pi, \pi]$, $i=1, \ldots, 6$. 

In order to use optimization algorithms which may leave the feasible set $\workspace_M$,
our goal is to define a {\em virtual robot} which has a solution for the backward transform
for any frame $F\in\frames$ and any configuration $s$. 
So we associate to our {\em original robot} a {\em virtual robot} with an 
additional {\em virtual prismatic joint}
between joints 3 and 4, which has no joint limits. Any \wcp in $\mR^3$ is reachable then.
The variable of the virtual joint will be denoted $v$, the other
joints keep their  names giving a combined joint variable 
$\tilde \axis = (\axis_1, \axis_2, \axis_3, v, \axis_4, \axis_5, \axis_6)\in\mR^7$.
DH parameters of the virtual robot are shown in 
Table \ref{fig:DHparametersVirtual}.
\begin{table}
  \caption{DH parameters for virtual robot}
  \label{fig:DHparametersVirtual}
\centering
  \begin{tabular}{c|c|c|c|c||c}
        $i$ \   & \ $\theta_i$ \ & \ $d_i$ \   &  \  $a_i$ \ & \ $\alpha_i$ \ & \ type\\ 
	\hline
         1   & $\tilde q_1$ & 0   & 0  &  $\frac \pi 2$ & R\\
         2   & $\tilde q_2$ &  0 &   $l_{23}$ & 0 & R\\
         3   & $\tilde q_3$ &  0 & 0 & $-\frac \pi 2$ & R\\
         4   & 0 & $v$ & 0 & 0 & P\\
         5   & $\tilde q_4$ &  $l_{35}$ & 0 & $\frac \pi 2$ & R\\
         6   & $\tilde q_5$ &  0 & 0  & $-\frac \pi 2$ & R\\
         7   & $\tilde q_6$ &  0   & 0 & 0 & R
  \end{tabular}
\end{table}

\begin{figure}
\centering
\includegraphics[width=60mm]{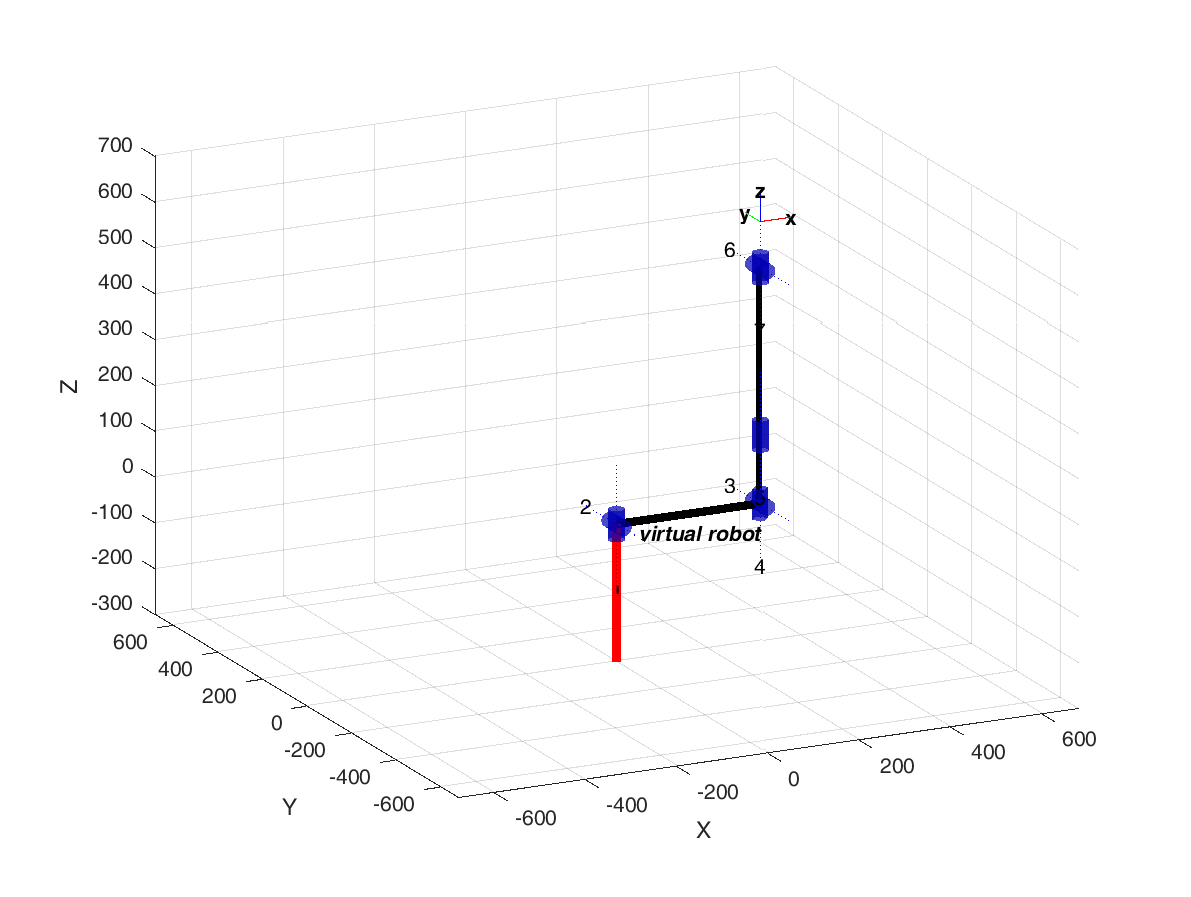}
  \caption{Reference position 
       $\tilde\axis=(0,0,0,150,0,0,0)$ for virtual robot}
  \label{fig:ReferenceVirtual}
\end{figure}

Sufficient conditions for our approach are stated as two assumptions:
{\bf Assumption 1 - Reachability of $\mR^3$:} The mapping of the 
original joints 
and the virtual joint to the \wcp position is surjective onto $\mR^3$.
{\bf Assumption 2 - Reachability of $\SO 3$} Joints 4,5,6 form a central wrist 
parametrizing all of $\SO 3$, i.e. 
the mapping $(-\pi, \pi]^3 \to \SO 3$, 
$(\axis_4, \axis_5, \axis_6)\mapsto 
\rotz {\axis_4} \cdot B_4\cdot\rotz {\axis_5} \cdot B_5\cdot\rotz {\axis_6} $ is surjective.

Both assumptions hold for standard 6R robots as considered here, for details see
\cite{WeissARK}.  
However we have introduced redundancy in our kinematics so we have to define a backward
transform giving unique results. The virtual robot backward transform sets the virtual 
joint to the smallest absolute value such that a solution exists. 
In our case this is the 
distance between the \wcp position and 
the workspace of the original robot
which is a hollow sphere for our robot
so calculations are simple. For algorithmic details see \cite{WeissARK} again.

\section{Formulation of the Optimization Problem}

We assume that in the unloading process of the box a linar motion in $z$ direction 
should be made. It is sufficient to consider a single grid point, and to build a loop
repeating the optimization over all grid points.

At a single grid point $P_0$ we are given a frame $F$ with $P_0$ the position part
and some orientation $Q$, as well as a target point $P_1$ which differs from
$P_0$ in the $z$ component by some $\Delta z$ only, 
and the same associated orientation $Q$. 
The corresponding frames $(Q,P_0)$ and $(Q, P_1)$ should be connected by a linear motion.
The motion geometry is parametrized by $\lambda\in[0,1]$, and discretized with step size
$h = \frac 1 N$ at interpolating points $\lambda_i = i \cdot h$, $i = 0, \ldots, N$, 
leading to frames $(Q, P_0 + (0,0,\lambda_i \cdot\Delta z)$. We want to exploit the rotation
around $z$ as an additional degree of freedom and introduce variables $\alpha_0$ and 
$\alpha_1$ for the rotations at $P_0$ and $P_1$, to be interpolated linearly by the robot
controller: 
$$F_i = (Q\cdot \rotz {\alpha_0 + \lambda_i\cdot (\alpha_1-\alpha_0)}, 
P_0 + (0,0,\lambda_i \cdot\Delta z).$$
The backward transform of the {\em virtual} robot gives 
$$\tilde \axis^{(i)} = \tilde \axis(F_i) = \tilde \axis(F(\alpha_0, \alpha_1)).$$ 
If 
the virtual joint is not needed, $v_i=0$, then the pose is admissible for the original
robot as well. So our objective is to minimize the absolute value of $\tilde v_i$, which
is expressed by 

\begin{eqnarray*}
	\min_{\alpha_0,\alpha_1\in]-\pi, \pi]} \frac 1 2 \sum_{i=0}^n v_i^2 
\end{eqnarray*}

The optimization is nonlinear as the backward transform is hidden in the computation of
$\tilde \axis$.
Note that the whole optimization can be generalized to more complicated motions as
long as the geometric algorithms of the robot controller are known.

\section{Numerical Results}

\begin{figure}
\centering
\includegraphics[height=60mm]{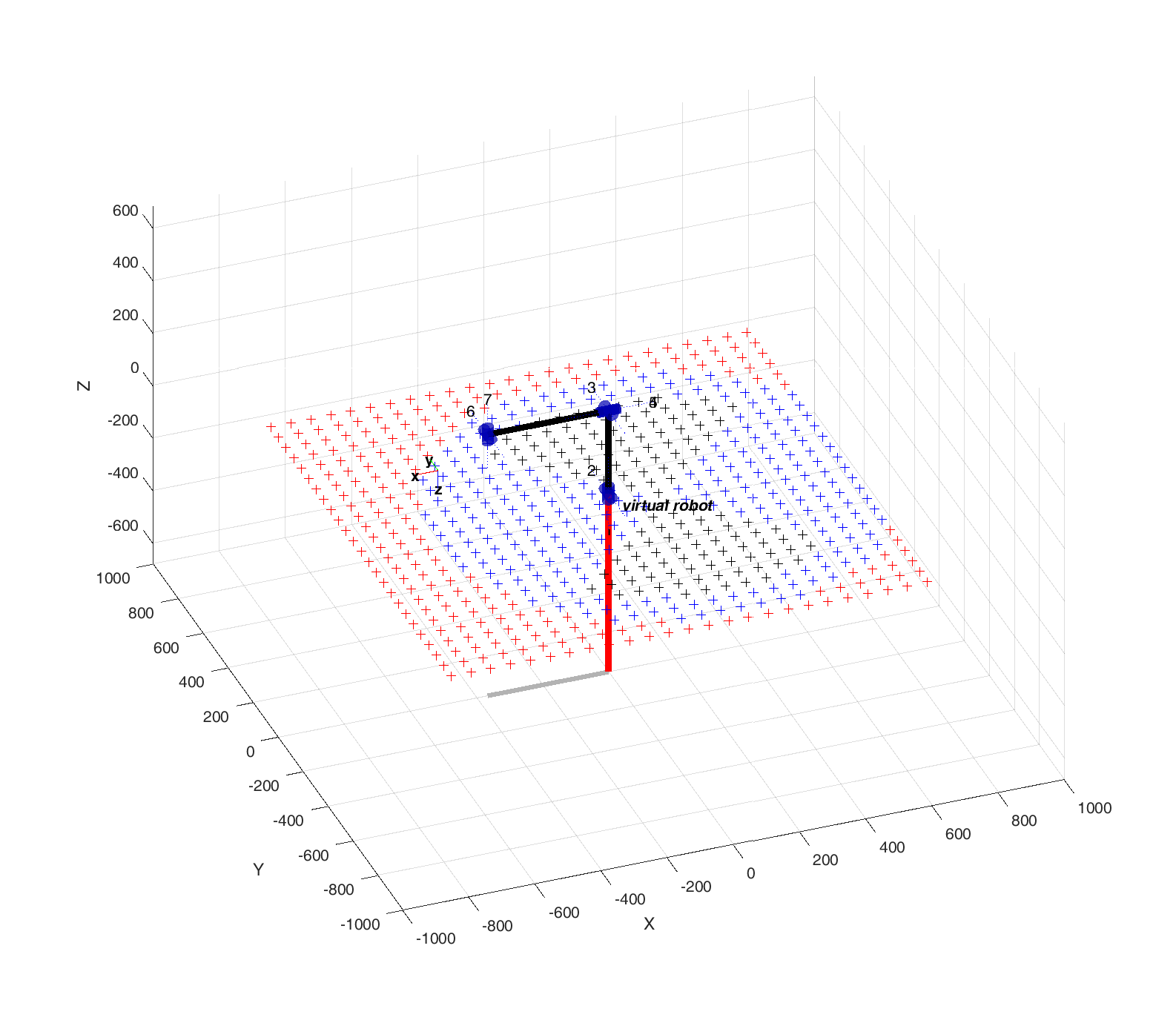}
\caption{Workspace violations}
\label{fig:Reachability2Reasons}
\end{figure}

We choose the tool geometry $(x,y,z) = (150,0,100)$ in our simulation. 
Figure \ref{fig:Reachability2Reasons} shows the two types of work space violations for a grid around the robot
in the $z$ plane: the grid points shoud be approached with the same orientation, and
some fixed configuration. 
Red points are outside of the robots work space, essentially unreachable for the \wcp
with the commanded orientation and configuration. Blue points are reachable for the \wcp,
but violate axis limits. Black points are reachable within the joint limits.

\begin{figure}
\centering
\includegraphics[height=60mm]{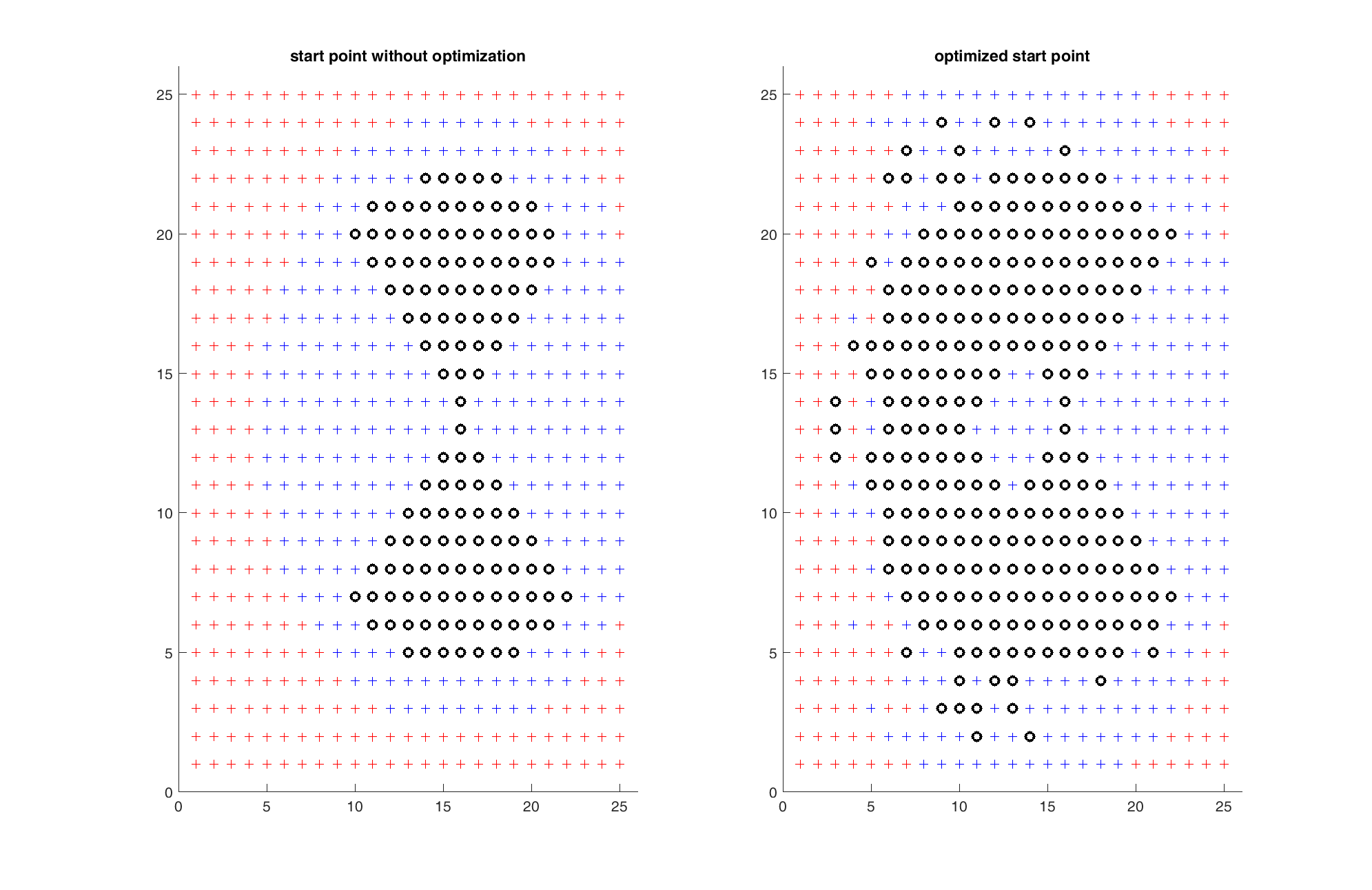}
\caption{Reachability for start point of linear motion}
\label{fig:Admissibility}
\end{figure}

Figure \ref{fig:Admissibility} 
shows the situation from the top before and after the optimization process: 
Instead of the same prescribed orientation around the symmetry axis of the objects 
at all grid points the optimizer chooses different orientations and achieves more grid 
points with admissible robot poses both for \wcp and axis limits, as can be seen from the
increase in black locations. 

\begin{figure}
\centering
\includegraphics[height=50mm]{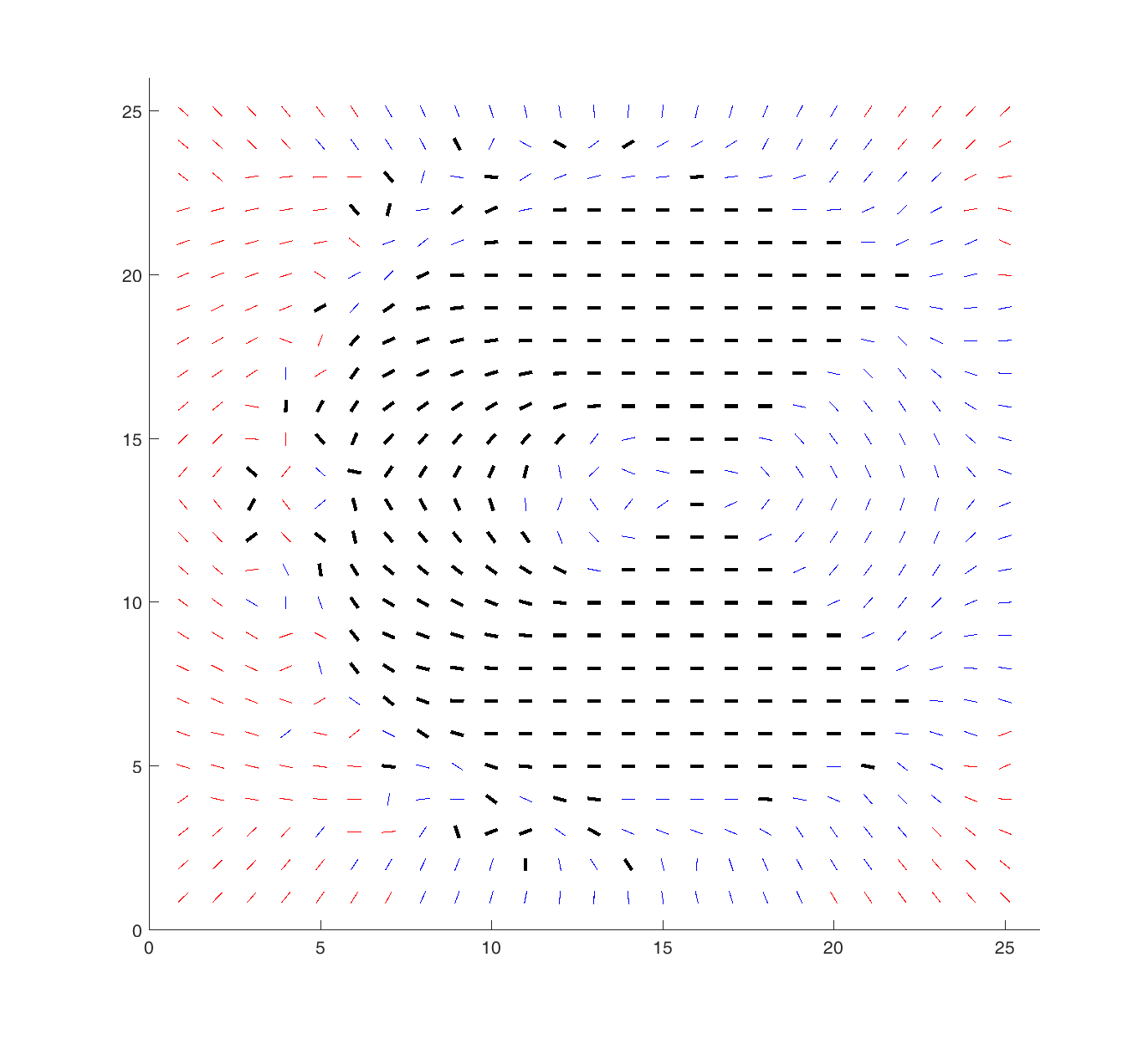}
\caption{Change in rotational degree of freedom by optimization}
\label{fig:ChangeInRotation}
\end{figure}

Figure \ref{fig:ChangeInRotation} shows a kind of direction field representation of
the chosen orientations: initially the rotational degree of freedom points to the right 
for all grid points. If this results in a legal pose the optimizer has no need to change 
anything. Otherwise the rotation around the symmetry axis is adjusted, resulting in 
some more legal (black) positions. For grid points where no legal pose could be found
the rotation gives an direction where either the distance of the \wcp to the workspace
or the axis limit violation decreases.

We have tested our optimization procedure with the solvers implemented in the MATLAB 
{\tt fmincon} command. We obtained optimal solutions both with the default 
interior point algorithm and the SQP algorithms. However, in many cases the SQP algorithm
required less iterations. Note that these algorithms require $C^2$ functions. The
approach described so far gives a continuous but not differentiable function. 
In \cite{WeissARK} a smoothing operation is explained similar to the
\equote{kinks} of \cite{Bertsekas:kinks} which overcomes this problem.

Computation time was below 10 sec on a standard laptop with $B_x\cdot B_y = 25\cdot 25 = 625$
grid points in the box.

\section{Conclusions}

We have shown how the virtual axis idea can be used for the optimization of
processes with redundant degrees of freedom. The next natural steps are: How to select 
optimal motions instead of just admissible ones? This seems difficult to incorporate 
in the objective function. How to use this approach for robots with more than 6 axes? Here
the choice of the correct configuration is a great challenge.


\begin{thebibliography}{6}
%

\bibitem{Bertsekas:kinks}
Bertsekas, D.P.: Nondifferentiable Optimization via Approximation. 
Math. Programming Study 3, pp. 1-25 (1975).


%
%
%
%
%

%
%

\bibitem{Craig}
 Craig, J.: Introduction to Robotics: Mechanics and Control.
 Addison Wesley, (1986)

%
%
%
%
%

%
%
%

\bibitem{Goppold}
Goppold, J.: Optimale Platzierung eines Objektes im Arbeitsraum eines Roboters,
Bachelor Thesis, Ostbayerische Technische Hochschule Regensburg (2016)
 
\bibitem{KUKA:StudentsWin}
KUKA Robot Group: Robots Play Board Games - Students Win Big, 
\url{https://www.youtube.com/watch?v=oCQPWv_ky2c}, (2016)

\bibitem{Leger:Angeles}
L{\'e}ger, J., Angeles, J.:
Off-line programming of six-axis robots for optimum
five-dimensional tasks.
Mechanism and Machine Theory 100, 155–-169  (2016)

\bibitem{Leontjevs}
Leontjevs, V., Flores, F.G., Lopes, J., Kecskemethy, A.:
Singularity Avoidance by Virtual Redundant Axis and its Application to 
Large Base Motion Compensation of Serial Robots
In: Proceedings of the RAAD 2012
21st International Workshop on Robotics in Alpe-Adria-Danube Region.
Naples (2012)

\bibitem{KUKASpezAgilus}
KUKA Roboter GmbH: KR AGILUS sixx Specification. Augsburg (2013)

\bibitem {Leger:Angeles}
L{\'e}ger, J., Angeles, J.:
Off-line programming of six-axis robots for optimum
five-dimensional tasks.
Mechanism and Machine Theory 100, 155–-169  (2016)

\bibitem{Merlet}
Merlet, J.P.: Jacobian, manipulability, condition number, and accuracy of parallel robots. J.
Mech. Des. 128(1), 199-–206 (2006)

\bibitem{Nocedal}
 Nocedal, J., Wright, S.J.:
 Numerical Optimization.
 Springer, New York (2006)

\bibitem{Reiter}
Reiter, A.: Ein Beitrag zur Singularit\"atsvermeidung bei
Industrierobotern durch Einf\"uhrung virtueller Achsen.
Master Thesis, Johannes Kepler University Linz (2015)

%
%
%
%

\bibitem{WeissARK}
Wei\ss, M.: Optimal Object Placement using a Virtual Axis, 
ARK 2018, 16th International Symposium on Advances in Robot Kinematics, 
Bologna, submitted (2018)

\bibitem{Yoshikawa}
Yoshikawa, T.: Manipulability of Robotic Mechanisms  
The International Journal of Robotics Research 4(2), pp. 3-9 (1985).




\end{thebibliography}
\end{document}